\documentclass{article}
\usepackage{spconf,amsmath,amssymb}
\usepackage{microtype,url}
\usepackage{graphicx,epsfig}
\usepackage[linesnumbered,boxed]{algorithm2e}
\usepackage[caption=false,font=footnotesize]{subfig}

\def\i{\boldsymbol{i}}
\def\j{\boldsymbol{j}}
\def\Z{\mathbb{Z}}

\title{Image Denoising using Optimally Weighted Bilateral Filters: \\ A Sure and Fast Approach}
\name{$^{\star}$Kunal N. Chaudhury\thanks{K.~N.~Chaudhury was partially supported by a Startup Grant provided by the Indian Institute of Science. K. Rithwik was supported by a fellowship from the Indian Academy of Sciences. Correspondence: kunal@ee.iisc.ernet.in.} \qquad $^{\S}$Kollipara Rithwik}

\address{$^{\star}$Department of Electrical Engineering, Indian Institute of Science, Bangalore, India \\
$^{\S}$Department of Electrical Engineering, Indian Institute of Technology, Hyderabad, India}

\begin{document}
\ninept

\maketitle

\begin{abstract}
The bilateral filter is known to be quite effective in denoising images corrupted with small dosages of additive Gaussian noise. 
The denoising performance of the filter, however, is known to degrade quickly with the increase in noise level. Several adaptations of the filter have been proposed in the literature to address this shortcoming, but often at a substantial computational overhead. 
In this paper, we report a simple pre-processing step that can substantially improve the denoising performance of the bilateral filter, at almost no additional cost. 
The modified filter is designed to be robust at large noise levels, and often tends to perform poorly below a certain noise threshold. 
To get the best of the original and the modified filter, we propose to combine them in a weighted fashion, where the weights are chosen to  minimize (a surrogate of) the oracle mean-squared-error (MSE). The optimally-weighted filter is thus guaranteed to perform better than either of the component filters in terms of the MSE, at all noise levels.
We also provide a fast algorithm for the weighted filtering.
Visual and quantitative denoising results on standard test images are reported which demonstrate that the improvement over the original filter is significant both visually and in terms of PSNR. 
Moreover, the denoising performance of the optimally-weighted bilateral filter is competitive with the computation-intensive non-local means filter.
\end{abstract}


\section{Introduction}
\label{sec:intro}

We consider the standard problem of denoising grayscale images that are corrupted with additive white Gaussian noise \cite{Yaroslavsky1985,Osher1992,Coifman1995}. 
In this setup, we are given the \textit{corrupted}  (or \textit{noisy}) image
\begin{equation}
\label{noise}
f(\i) = f_0(\i) + \sigma \cdot w_{\i} \qquad (\i \in I),
\end{equation}
where $I$ is some finite rectangular domain of $\Z^2$, $\{f_0(\i) : \i \in I\}$ is the unknown \textit{clean} image, $\{w_{\i} : \i \in I\}$ are independent and distributed as $\mathcal{N}(0,1)$, and $\sigma$ is the noise level.

The goal is find a \textit{denoised} estimate $\hat{f}(\i)$ of the clean image from the corrupted samples. The denoised image should visually resemble the clean image. One way to quantify the resemblance is using the mean-squared-error (MSE) which is defined to be
\begin{equation}
\label{MSE}
\text{MSE}= \frac{1}{|I|} \sum_{\i \in I} \big(\hat{f}(\i) - f_0(\i) \big)^2.
\end{equation}
Later in the paper, we will use the peak signal-to-noise ratio (PSNR) which is given by  $\text{PSNR}=10 \log_{10}(255^2/\text{MSE})$. The image denoising problem has been extensively studied and a survey of even a fraction of this literature is beyond the scope of this paper. We instead refer the interested reader to \cite{Yaroslavsky1985} - \cite{Milanfar2013} and the references therein.

Our present interest is in the image denoising applications of the edge-preserving bilateral filter \cite{Tomasi1998,Aleksic2006,Liu2006}.
The denoised image in this case is set to be 
\begin{equation}
\label{BF}
 \hat{f}_1(\i)=  \frac{\sum_{\j \in \Omega} g_{\sigma_s}(\j) \  g_{\sigma_r}\left(f(\i-\j)-f(\i)\right) \ f(\i-\j)}{\sum_{\j \in \Omega} g_{\sigma_s}(\j)  \ g_{\sigma_r}\left(f(\i-\j)-f(\i)\right) }.
\end{equation}
where  
\begin{equation}
\label{kernel}
g_{\sigma_s}(\i) = \exp\left(- \frac{\lVert \i \rVert^2}{2\sigma_s^2}\right)  \quad \text{and}  \quad  g_{\sigma_r}(t) = \exp\left(- \frac{t^2}{2\sigma_r^2}\right).
\end{equation}
The Gaussian kernels in \eqref{kernel} are respectively referred to as the \textit{spatial} and \textit{range} kernels. In practice, the domain $\Omega$ is restricted to some neighbourhood of the origin. Typically, $\Omega$ is a square neighbourhood, $\Omega=[-W,W] \times [-W,W]$, where $W=3\sigma_s$ \cite{Tomasi1998}. We refer the reader to \cite{Tomasi1998,Book2009} for a detailed exposition on the working and in particular the edge-preserving property of the filter.

The bilateral filter has received renewed attention in the image processing community in the context of image denoising \cite{Knaus2014,Morel2014}. 
It is well-known that, while the filter is quite effective in removing modest levels of additive noise, its denoising performance drops at large noise levels \cite{Book2009}. It was demonstrated in \cite{Buades2005,Awate2006} that a patch-based extension of the filter can be used to bring the denoising performance of the filter at par with state-of-the-art methods. These, and other advanced patch-based methods \cite{Kervrann2006,KSVD,BM3D}, are however much more computation-intensive than the bilateral filter. 

\subsection{Present Contribution}
 
In this work, we present a couple of ideas for improving the denoising performance of the classical bilateral filter, and give fast algorithms for the same. The contributions (and the organization) of the paper are as follows:

$\bullet$ In Section \ref{sec:RBF}, we demonstrate how the denoising performance of the bilateral filter can be improved at large noise levels (at almost no additional cost) by incorporating a simple pre-processing step into the framework.  
 
$\bullet$ The modified filter is designed to be robust at large noise levels, and tends to perform poorly below a certain noise threshold.  To get the best of the either filters at all noise levels, we propose to optimally combine them in a weighted fashion in Section \ref{sec:OWF}. The optimality is in terms of a certain surrogate of the mean-squared-error (MSE) known as Stein's unbiased risk estimate (SURE) \cite{Stein1981}. This MSE-estimate is known to be quite accurate in practice \cite{Luisier2007,Peng2010,Kisan2012} and, as a result, the combined filter is almost always guaranteed to provide a lower MSE than the original filter. 
 
$\bullet$ Following \cite{Chaudhury2011,Chaudhury2013}, we present an approximation for the proposed filtering in Section \ref{sec:fastSURE} that has a fast implementation.  We derive the SURE estimator for this approximation, and demonstrate how it can be efficiently computed in the process of approximating the bilateral filter. The overall cost of computing the estimator and the optimally-weighted filters is about twice the cost of computing a single bilateral filtering using the fast algorithm in \cite{Chaudhury2011}. 

We provide both visual and quantitative denoising results on standard test images in Section \ref{sec:experiments} which demonstrate that the improvement over the original filter is significant both visually and in terms of PSNR.

\section{Robust Bilateral Filter}
\label{sec:RBF}

Notice that the range filter in \eqref{BF} operates on the noisy samples. In other words, the corrupted image is used not just for the averaging but also to control the blurring via the range filter. What if the range filter could directly operate on the clean image? It can be verified that the denoising image obtained using this ``oracle'' filter is visibly better and has higher PSNR, which is not surprising. Of course, we do not have access to the clean image in practice, and thus the oracle bilateral filter cannot be realized. Nevertheless, one could consider some form of proxy for the clean image. Our proposal is simply to use the box-filtered version of the noisy image as a proxy. 
In particular, we propose the following robust bilateral filter (RBF):
\begin{equation}
\label{RBF}
 \hat{f}_2(\i)=  \frac{\sum_{\j \in \Omega} g_{\sigma_s}(\j) \  g_{\sigma_r}(\bar{f}(\i-\j)- \bar{f}(\i)) \ f(\i-\j)}{\sum_{\j \in \Omega} g_{\sigma_s}(\j)  \  g_{\sigma_r}(\bar{f}(\i-\j)-\bar{f}(\i)) }.
\end{equation}
where
\begin{equation}
\label{box}
 \bar{f}(\i)= \frac{1}{(2L+1)^2} \sum_{\j \in \{-L,L\}^2} \!\!\! f(\i-\j).
\end{equation}
The amount of smoothing induced by the box filter is controlled by $L$. 
%When $L$ is very small, $ \bar{f}(\i) \approx f(i)$, and \eqref{IBF} behaves as \eqref{BF}. At the other other end, the image structures are over-smoothed when $L$ is large %and this makes $\bar{f}(\i)$ a bad proxy for the original image. 
We performed exhaustive simulations in this direction. The simulation  results suggest that $L=1$ ($3 \times 3$ blur) is optimal for most settings. A possible way to explain this is that this small blur is able to suppress the noise without excessively blurring the image features.
The denoising results obtained using the standard and the robust filter a couple of test images are shown in Table \ref{table1}. 
We note that the filters have been independently tuned with respect to $\sigma_s$ and $\sigma_r$ to get the best PSNR. 
These results (and the results on other test images that are not shown here) suggest that the robust filter starts to perform better beyond a certain noise level depending on the type of image. This is not surprising since, at low noise levels, the box filtering introduces more blurring than noise suppression which brings down the overall signal-to-noise ratio. Indeed, the corrupted image is already a good proxy for the clean image when the noise floor is small. On the other hand, notice that the improvement in PSNR is often as large as $10$ dB at large noise levels. 

\section{Optimally Weighted Bilateral Filters}
\label{sec:OWF}

The results in Table \ref{table1} suggest that we could possibly perform better denoising by combining \eqref{BF} and \eqref{RBF}. A particularly simple idea would be to take a linear combination of these estimates. That is, we could set the denoised image to be
\begin{equation}
\label{WBF}
\hat{f}(\i) = \theta_1 \hat{f}_1(\i) +  \theta_2 \hat{f}_2(\i) \qquad (\i \in I).
\end{equation}
Notice that this includes \eqref{BF} and \eqref{RBF} as special cases. 
%Moreover, we note that \eqref{WBF} can be seen as a bilateral filter whose range kernel is a weighted sum of the kernels in \eqref{BF}and \eqref{RBF}. However, we do not gain %anything from this perspective in terms of computations, and thus we prefer to write it as in \eqref{WBF}.

This bring us to the question of setting the weights   $\theta_1$ and $\theta_2$. This can hypothetically be done by minimizing the MSE between \eqref{WBF} and the clean image. However, we do not have access to the clean image. This is precisely where the classical Stein's unbiased risk estimate (SURE) \cite{Stein1981} is useful. This is given by
\begin{equation}
\label{sure}
\text{SURE} = \frac{1}{| I |} \sum_{\i \in I} \big(\hat{f}(\i) - f(\i) \big)^2 - \sigma^2 + \frac{2\sigma^2}{|I|} \sum_{\i \in I} \frac{\partial \hat{f}(\i)}{\partial f(\i)}.
\end{equation}
SURE has the property that its expected value equals that of \eqref{MSE} for the Gaussian noise model in \eqref{noise} \cite{Stein1981}.
This makes it an useful surrogate for the MSE, which can be computed without the knowledge of the clean image.
We note that the idea of taking linear (or affine) combinations of estimators and tuning the weights to get the optimal SURE has been tried in different contexts in the literature; e.g., see \cite{Luisier2007}. Note that the added computation required for SURE are the computation of the partial derivatives in \eqref{sure}. 

We propose to find the optimal weights in \eqref{WBF} by minimizing SURE. In particular, on substituting \eqref{WBF} in \eqref{sure}, we see that SURE is quadratic in $\theta_1$ and $\theta_2$. Thus, by convexity, a necessary and sufficient condition for optimality is that that the gradient of \eqref{sure} must vanish at the optimal weights. In particular, it can be verified that the resulting gradient equations can be written as $\mathbf{A} \boldsymbol{\theta}^{\#} = \boldsymbol{b}$, where
\begin{equation}
\label{A}
\mathbf{A} = \begin{pmatrix}
  \sum_{\i \in I} \hat{f}_1(\i)^2 & \hspace{2mm} \sum_{\i \in I} \hat{f}_1(\i)\hat{f}_2(\i) \\[6pt]
 \sum_{\i \in I} \hat{f}_1(\i)\hat{f}_2(\i)  &  \hspace{2mm} \sum_{\i \in I} \hat{f}_2(\i)^2
 \end{pmatrix},
\end{equation}
and
\begin{equation}
\label{b}
\boldsymbol{b} = \begin{pmatrix}
  \sum_{\i \in I} f(\i) \hat{f}_1(\i) - \sigma^2 \sum_{\i \in I} \partial_{\i} \hat{f}_1(\i)  \\[6pt] 
   \sum_{\i \in I} f(\i) \hat{f}_2(\i) - \sigma^2 \sum_{\i \in I} \partial_{\i} \hat{f}_2(\i)  \end{pmatrix},
\end{equation}
and $\boldsymbol{\theta}^{\#} =(\theta^{\#}_1,  \ \theta^{\#}_2)$ are the optimal weights. To simply notations, we will henceforth use the shorthand $\partial_{\i}$ in place of the  operator $\partial/ \partial f(\i)$ appearing in \eqref{sure}. In summary, by computing $\mathbf{A}$ and $\boldsymbol{b}$, and solving the  $2 \times 2$ linear equation $\mathbf{A} \boldsymbol{\theta}^{\#} = \boldsymbol{b}$, we get the optimal weights. The weights are then plugged into \eqref{WBF} to get the final denoised image.

\begin{table}
\setlength{\tabcolsep}{4pt}
\caption{Comparison of the standard bilateral filter (SBF) and the robust bilateral filter (RBF) at $\sigma = 10, 20, 30, 40, 50, 60$. For a fixed image and noise level, we tuned $\sigma_s$ and $\sigma_r$  to individually optimize the PSNR obtained using both methods.}  
\vspace{2mm}
\centering 
\begin{tabular}{l  c rrrrrr}  
\hline 

Image & Filter &\multicolumn{6}{c}{PSNR (dB)} \\

\hline

% House
& SBF    &\textbf{33.76}		&29.88		&25.48	&21.44	&18.27	&15.83  \\
\raisebox{1ex}{\textit{House}} 
& RBF    &33.15		&\textbf{31.35}		&\textbf{29.85}	&\textbf{28.33}	&\textbf{27.23}	&\textbf{26.27}   \\

\hline
% Peppers
& SBF    &\textbf{32.94}		&28.97		&24.88	&20.86	&17.89	&15.56  \\
\raisebox{1ex}{\textit{Peppers}} 
& RBF    &31.30	&\textbf{29.73}	&\textbf{27.92}	&\textbf{26.31}	&\textbf{25.17}	&\textbf{24.27}  \\

\hline 

\end{tabular}
\label{table1}
\end{table}

\section{Fast and SURE Implementation}
\label{sec:fastSURE}

The cost of computing \eqref{BF} and \eqref{RBF} is clearly $O(\sigma_s^2)$ per pixel, since the support of the spatial filter is proportional to $\sigma_s^2$. Moreover, we are also required to compute the derivatives for SURE which would further add to the cost. 
We now explain how we can compute \eqref{BF}, \eqref{RBF}, and the derivatives $\partial_{\i} \hat{f}_1(\i)$ and $\partial_{\i} \hat{f}_2(\i)$ using the fast algorithm in \cite{Chaudhury2011,Chaudhury2013}. It was observed here that, for sufficiently large $N$, we can accurately approximate the range kernel in \eqref{kernel} using
\begin{equation}
\label{shiftability}
\left[ \cos\left( \frac{t}{\sigma_r \sqrt{N}} \right) \right]^N= \sum_{n=0}^N c_n  \exp \left(\imath \omega_n t \right),
\end{equation}
where $\imath$ denotes $\sqrt{-1}$, 
\begin{equation}
\label{coeff}
c_n = \frac{1}{2^N}\binom{N}{n}, \quad \text{and} \quad \omega_n=\frac{(2n-N)}{\sigma_r \sqrt{N}}.
\end{equation}
Plugging \eqref{shiftability} into \eqref{BF}, using the multiplication-addition property of exponentials, and exchanging the summations, the numerator of \eqref{BF} 
can be expressed as
\begin{equation*}
P(\i)=\sum_{n=0}^N  c_n  H_n(\i)  \bar{F}_n(\i),
\end{equation*}
where 
$H_n(\i) = \exp (-\imath \omega_n f(\i)), F_n(\i) = H^{\star}_n(\i) f(\i)$, and $\bar{F}_n(\i)$ denotes the Gaussian filtering of $F_n$:
\begin{equation}
\label{gauss}
\bar{F}_n(\i) = (F_n \ast g_{\sigma_s})(\i)= \sum_{\j \in \Omega} g_{\sigma_s}(\j) F_n(\i-\j)
\end{equation}
Here and later, we use $z^{\star}$ to denote the complex conjugate of $z$. Similarly,  the denominator of \eqref{BF} can be expressed as
\begin{equation*}
Q(\i) =  \sum_{n=0}^N  c_n  H_n(\i) \bar{G}_n(\i),
\end{equation*}
where $G_n(\i) = H^{\star}_n(\i) $ and $\bar{G}_n(\i)$ is as defined in \eqref{gauss}. In summary, we can approximate \eqref{BF} using
\begin{equation}
\label{approxBF}
\hat{f}_1(\i) = \frac{P(\i)}{Q(\i)}.
\end{equation}
We note that the same notation has been used for both  \eqref{BF} and its approximation \eqref{approxBF}. The computation of \eqref{approxBF} is clearly dominated by the computation of a series of Gaussian filtering. Now, each of these filtering can implemented in constant-time (for any arbitrary $\sigma_s$) using separability and recursions \cite{Deriche1993}. We have thus been able to cut down the per-pixel complexity from $O(\sigma_s^2)$ to $O(1)$ using \eqref{approxBF}.
On the other hand, note that we can write
\begin{equation*}
\partial_{\i} \hat{f}_1(\i)  = \frac{1}{Q(\i)} \left(\partial_{\i}  P(\i) -\hat{f}_1(\i) \ \partial_{\i}  Q(\i) \right).
\end{equation*}
After some calculation and simplification, we can verify that 
\begin{equation}
\label{delP}
\partial_{\i}  P(\i) = g_{\sigma_s}(0) - \imath \sum_{n=0}^N c_n \omega_n H_n(\i) \bar{F}_n(\i),
\end{equation}
and
\begin{equation}
\label{delQ}
\partial_{\i}  Q(\i) = - \imath \sum_{n=0}^N c_n \omega_n H_n(\i) \bar{G}_n(\i).
\end{equation}
To arrive at the above formulas, we have use the fact that  the sum of $(c_n)$ is $1$ and that of $(c_n \omega_n)$ is $0$.
These identities are obtained by evaluating the both sides of \eqref{shiftability} and its derivative at $t=0$. 

%We note that $g_{\sigma_s}(0)=1$ from \eqref{kernel}. However, in practice, some form of discretization is used for the continuous Gaussian kernel, which is followed by a %normalization to ensure that the weights sum to unity \cite{Yaroslavsky1985}. In \eqref{delP} and \eqref{delQ}, we use $ g_{\sigma_s}(0)$ to denote the weight assigned to the %discrete kernel at the origin (which will generally be different from unity). 

The important point here is that some of the quantities that are used for computing \eqref{approxBF} are reused to compute the
partial derivatives in \eqref{delP} and \eqref{delQ}. The steps of the computation are summarized in Algorithm \ref{algo1}. It is clear that the main computations are the Gaussian filtering in steps $12$ and $13$. That is, the overall cost is dominated by the cost of computing $2(N+1)$ Gaussian filtering. Notice that we have recursively computed the binomial coefficients in \eqref{coeff}, and we have introduced temporary variables to cut down some of the redundant computations. 

\begin{algorithm}
\KwData{Noisy image $f(\i)$, and parameters $\sigma_s,\sigma_r,N$.}
\KwResult{$\hat{f}_1(\i), \partial_{\i}  P(\i)$, and $\partial_{\i}  Q(\i)$.}
$P(\i)=0$\;
$Q(\i)=0$\;
$\partial_{\i} P(\i)=0$\;
$\partial_{\i} Q(\i)=0$\;
$c=2^{-N}, \ \nu=1/(\sigma_r \sqrt{N})$\;
\For{$n=0,1,\ldots,N$}{
$c=  c  (N-n)/ (n+1)$\;
$\omega =(2n-N)\nu$\;
$H(\i) = \exp (-\imath \omega f(\i))$\;
$G(\i) = H^{\star}(\i)$\;
$F(\i) =  G(\i) f(\i)$\;
$B(\i) =  c \ H(\i) (F \ast g_{\sigma_s})(\i)$\;
$C(\i) =  c  \ H(\i) (G \ast g_{\sigma_s})(\i)$\;
$P(\i) = P(\i) + B(\i)$\;
$Q(\i) = Q(\i) + C(\i)$\;
$\partial_{\i} P(\i)=\partial_{\i} P(\i)+ \omega \  B(\i)$\;
$\partial_{\i} Q(\i)=\partial_{\i} Q(\i)+ \omega \ C(\i)$\;
}
$\hat{f}_1(\i)= P(\i)/Q(\i)$\;
$\partial_{\i} P(\i) =  g_{\sigma_s}(0) - \imath \ \partial_{\i} P(\i)$\;
$\partial_{\i} Q(\i) =   - \imath \ \partial_{\i} Q(\i)$\;
\caption{Computation of \eqref{approxBF}, \eqref{delP}, and \eqref{delQ}.}
\label{algo1}
\end{algorithm}

We can proceed similarly as above to approximate \eqref{RBF} and compute the associated derivatives. In particular, we can approximate \eqref{RBF} as 
$\hat{f}_2(\i) = R(\i)/S(\i)$, where
\begin{equation}
\label{RS}
R(\i)=\sum_{n=0}^N  c_n  U_n(\i)  \bar{V}_n(\i) \  \text{ and } \ S(\i) =  \sum_{n=0}^N  c_n  U_n(\i) \bar{W}_n(\i).
\end{equation}
Here $U_n(\i) = \exp (-\imath \omega_n \bar{f}(\i))$, $W_n(\i) = U^{\star}_n(\i)$, and $V_n(\i) = W_n(\i)f(\i)$.
On the other hand, after some calculation, we find that
\begin{equation*}
\partial_{\i} \hat{f}_2(\i)  = \frac{1}{S(\i)} \left(\partial_{\i}  R(\i) -\hat{f}_2(\i) \ \partial_{\i}  S(\i) \right),
\end{equation*}
where 
\begin{equation}
\label{delR}
\partial_{\i}  R(\i) = g_{\sigma_s}(0) - \frac{\imath}{(2L+1)^2} \sum_{n=0}^N c_n \omega_n U_n(\i) \bar{V}_n(\i),
\end{equation}
and
\begin{equation}
\label{delS}
\partial_{\i}  S(\i) = - \frac{\imath}{(2L+1)^2}  \sum_{n=0}^N c_n \omega_n U_n(\i) \bar{W}_n(\i).
\end{equation}
Notice that \eqref{delR} and \eqref{delS} are respectively identical to \eqref{delP} and \eqref{delQ} except for the additional $1/(2L+1)^2$ term. This term appears owing to the fact that both $R(\i)$ and $S(\i)$ are functions of $\bar{f}(i)$. In particular, we get this term from the application of the chain rule and the fact that  $\partial_{\i} \bar{f}(\i) = 1/(2L+1)^2$. Needless to say, the algorithm for computing the quantities in \eqref{RS}, \eqref{delR}, and \eqref{delS} is similar to Algorithm \ref{algo1}. The complete Matlab implementation 
can be found here \cite{MatlabFileExchange}.

\section{Experiments}
\label{sec:experiments}

\begin{table}
\setlength{\tabcolsep}{2.5pt}
\caption{Comparison of the Standard Bilateral Filter (SBF), the Robust Bilateral Filter (RBF), the Weighted Bilateral Filter (WBF), and the Non-Local Means (NLM) filter \cite{Buades2005} at noise levels $\sigma = 10, 15, 20, 25, 30,40, 50, 60$. We used the following test images \cite{USCdatabase}: \textit{Boat} (\textit{B}), \textit{Lena} (\textit{L}), \textit{House} (\textit{H}), \textit{Peppers} (\textit{P}), and \textit{Cameraman} (\textit{C}).
%We note that, for a fixed $\sigma_s$ and $\sigma_r$, the PNSR of WBF is automatically higher than that of SBF and RBF by construction; hence, we do not report this case.
}  
\vspace{2mm}
\centering 

\begin{tabular}{l  c rrrrrrrrrrr}  
\hline 

Image & Filter &\multicolumn{8}{c}{PSNR (dB)} \\

\hline

% Boats
& SBF    &32.02	&29.87	&28.44	&26.84	&24.86	&21.21	&18.20	&15.76  \\
& RBF    &29.95	&29.51	&28.90	&28.17	&27.46	&26.42	&25.56	&24.84  \\
\raisebox{0.4ex}{\textit{B}} 
& WBF    &\textbf{32.31}	&\textbf{30.44}	&\textbf{29.27}	&\textbf{28.33}	&\textbf{27.58}	&\textbf{26.50}	&\textbf{25.60}	&\textbf{24.86} \\
& NLM    &31.93	&29.93	&28.57	&27.6	&26.90	&25.68	&24.58	&23.84  \\

\hline

% Lena
& SBF    &33.61	&31.61	&30.07	&27.97	&25.59	&21.60	&18.39	&15.83  \\
& RBF    &33.30	&32.48	&31.49	&30.59	&29.84	&28.60	&27.63	&26.76   \\
\raisebox{0.4ex}{\textit{L}} 
& WBF    &\textbf{34.31}	&\textbf{32.75}	&\textbf{31.56}	&\textbf{30.64}	&\textbf{29.87}	&\textbf{28.62}	&\textbf{27.64}	&\textbf{26.76}  \\
& NLM    &34.06	&32.33	&30.99	&29.89	&29.07	&27.74	&26.72	&25.85  \\

\hline
% House
& SBF    &33.76	&31.54	&29.88	&27.77	&25.48	&21.44	&18.27	&15.83  \\
& RBF    &33.15	&32.34	&31.35	&30.56	&29.85	&28.33	&27.23	&26.27   \\
\raisebox{0.4ex}{\textit{H}} 
& WBF    &34.40	&32.66	&31.53	&\textbf{30.63}	&\textbf{29.90}	&\textbf{28.35}	&\textbf{27.24}	&\textbf{26.27}  \\
& NLM    &\textbf{34.63}	&\textbf{33.00}	&\textbf{31.63}	&30.56	&29.34	&27.69	&26.36	&25.00  \\

\hline
% Peppers
& SBF    &32.94	&30.71	&28.97	&27.01	&24.88	&20.86	&17.89	&15.56  \\
& RBF    &31.30	&30.60	&29.73	&28.79	&27.92	&26.31	&25.17	&24.27  \\
\raisebox{0.4ex}{\textit{P}} 
& WBF    &\textbf{33.38}	&\textbf{31.29}	&\textbf{29.95}	&\textbf{28.88}	&\textbf{27.97}	&\textbf{26.33}	&\textbf{25.20}	&\textbf{24.30} \\
& NLM    &32.91	&30.71	&29.23	&28.06	&27.14	&25.51	&24.40	&23.35  \\

\hline

% Cameraman
& SBF     &32.66	&30.20	&28.55	&26.80	&24.77	&21.16	&18.12	&15.59  \\ 
& RBF    &27.57	&27.34	&26.98	&26.45	&25.87	&25.00	&24.28	&23.59   \\
\raisebox{0.4ex}{\textit{C}} 
& WBF    &\textbf{32.69}	&\textbf{30.28}	&\textbf{28.61}	&\textbf{27.25}	&\textbf{26.36}	&\textbf{25.28}	&\textbf{24.41}	&\textbf{23.66}  \\
& NLM    &32.61	&30.00	&28.57	&27.70	&27.01	&25.39	&24.28	&23.22  \\

\hline

\end{tabular}
\label{table2}
\end{table}

We now present some denoising results on standard test images. Our objective is to compare the denoising results obtained using the proposed modification with the standard bilateral filter, both visually and in terms of PSNR. In this work, we assume that $\sigma$ is provided; this has to be estimated in practice from the noisy image. 
In this regard, we note that it has been demonstrated in \cite{Peng2010,Kisan2012} that the SURE is quite robust to standard data-based estimates of $\sigma$.  
We remark that one can optimize the SURE for the proposed filter with respect to $\sigma_s$ and $\sigma_r$ as done in \cite{Peng2010,Kisan2012}; however, we do not investigate this possibility in this paper.  

\begin{table}[!htb]
\caption{Comparison of the average run times of the direct and the fast implementation of the weighted bilateral filter for different $(\sigma_s,\sigma_r)$. We used  \textit{Barbara} ($512 \times 512$) for the comparison and set $\sigma=20$. The implementation was done using Matlab on an Intel quad-core 2.7 GHz machine with 8 GB memory.
For both implementations, we took the support of the Gaussian to be $W=3\sigma_s$.
 }  
\vspace{2mm}
\centering 
\begin{tabular}{l*{6}{c}r}
\hline
     &       $(2,15)$  & (4,20) & $(3,25)$  & $(5,30)$ & $(3,35)$  & $(4,40)$ \\
\hline
Direct          & 32s         & 120s    & 70s     & 185s    & 74s        & 125s   \\
Fast            & 1.2s         & 0.8s    &  0.7s   & 0.6s    & 0.5s     & 0.6s  \\
\hline
\end{tabular}
\label{table3}
\end{table}

\begin{figure}
\centering
\subfloat[\textit{Cameraman} ($256 \times 256$).]{\includegraphics[width=0.5\linewidth]{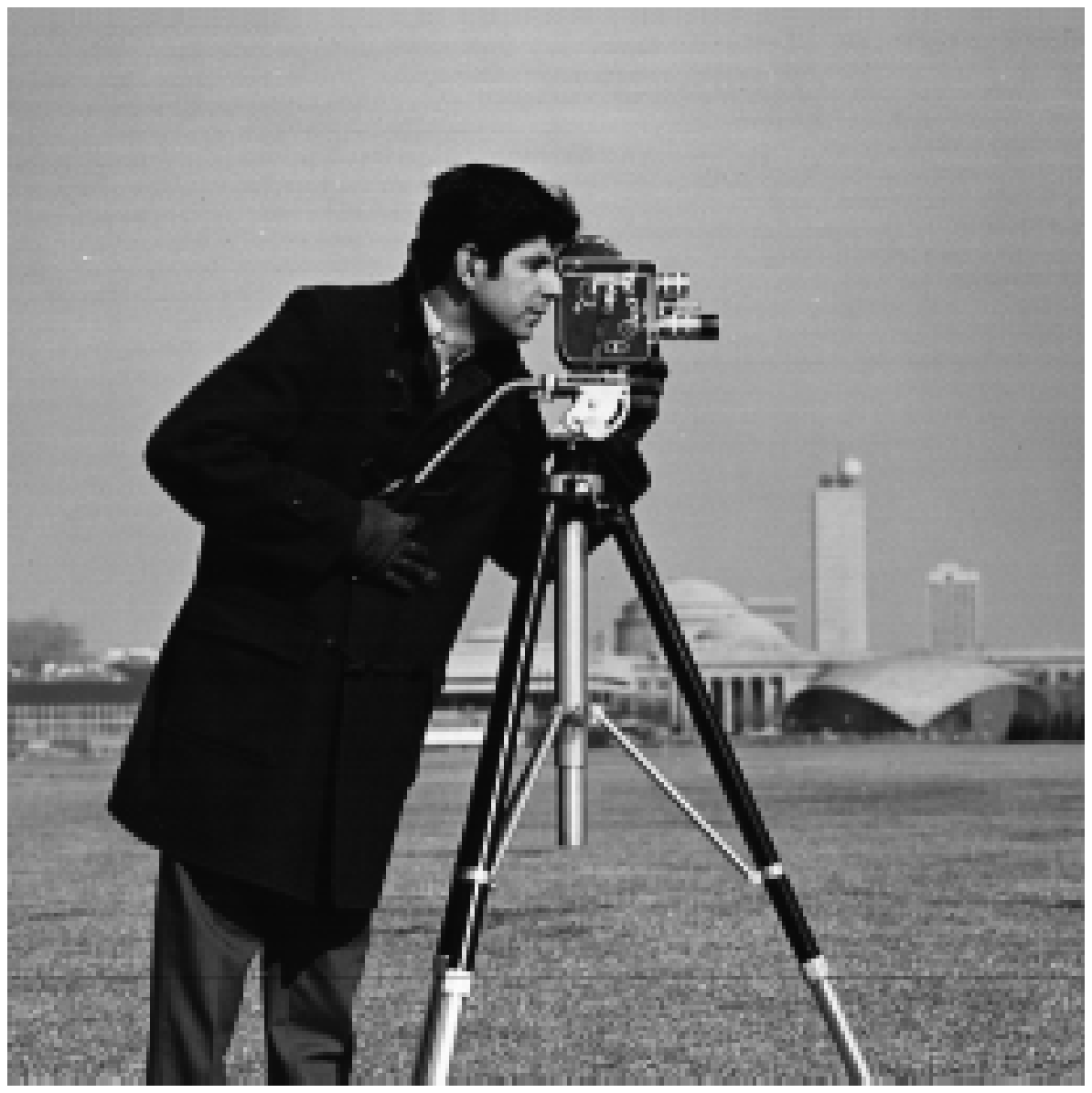}}  
\subfloat[Corrupted: $18.61$ dB ($\sigma = 30$).]{\includegraphics[width=0.5\linewidth]{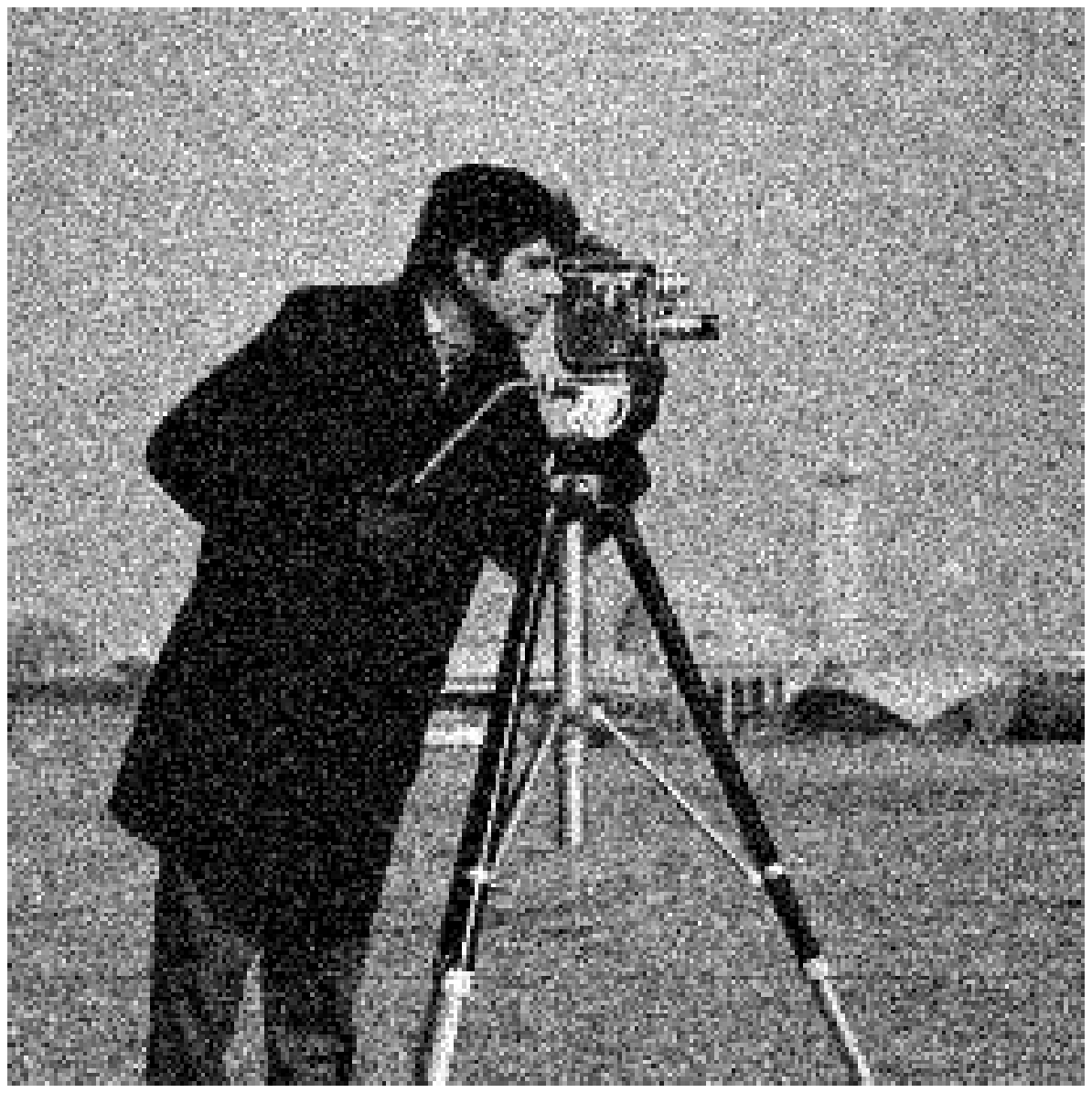}}  \\ \vspace{-1em}
\subfloat[SBF: $24.76$ dB; ($2, 45$).]{\includegraphics[width=0.5\linewidth]{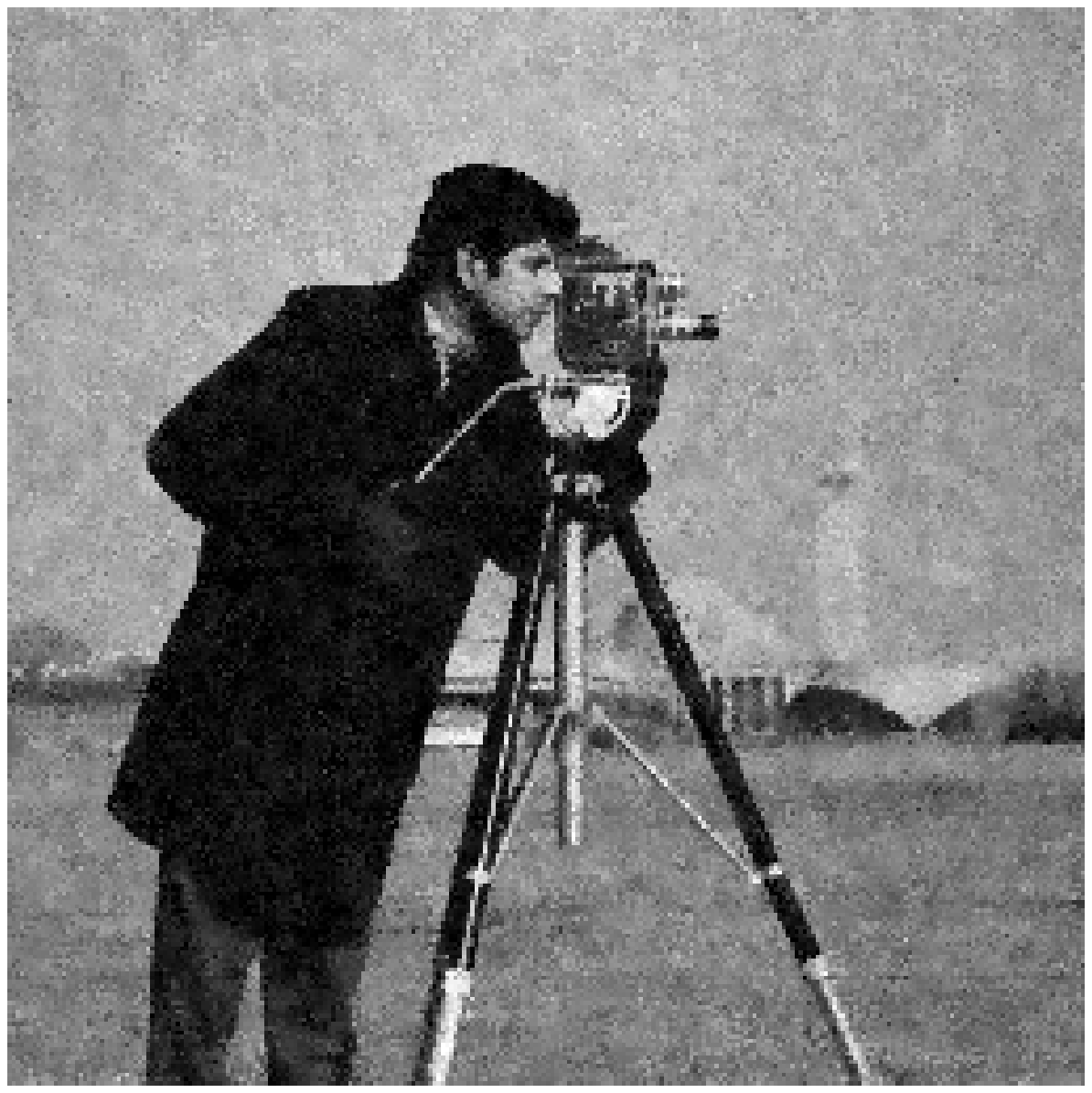}} 
\subfloat[WBF: $\bf{26.38}$ dB; ($4, 20$).]{\includegraphics[width=0.5\linewidth]{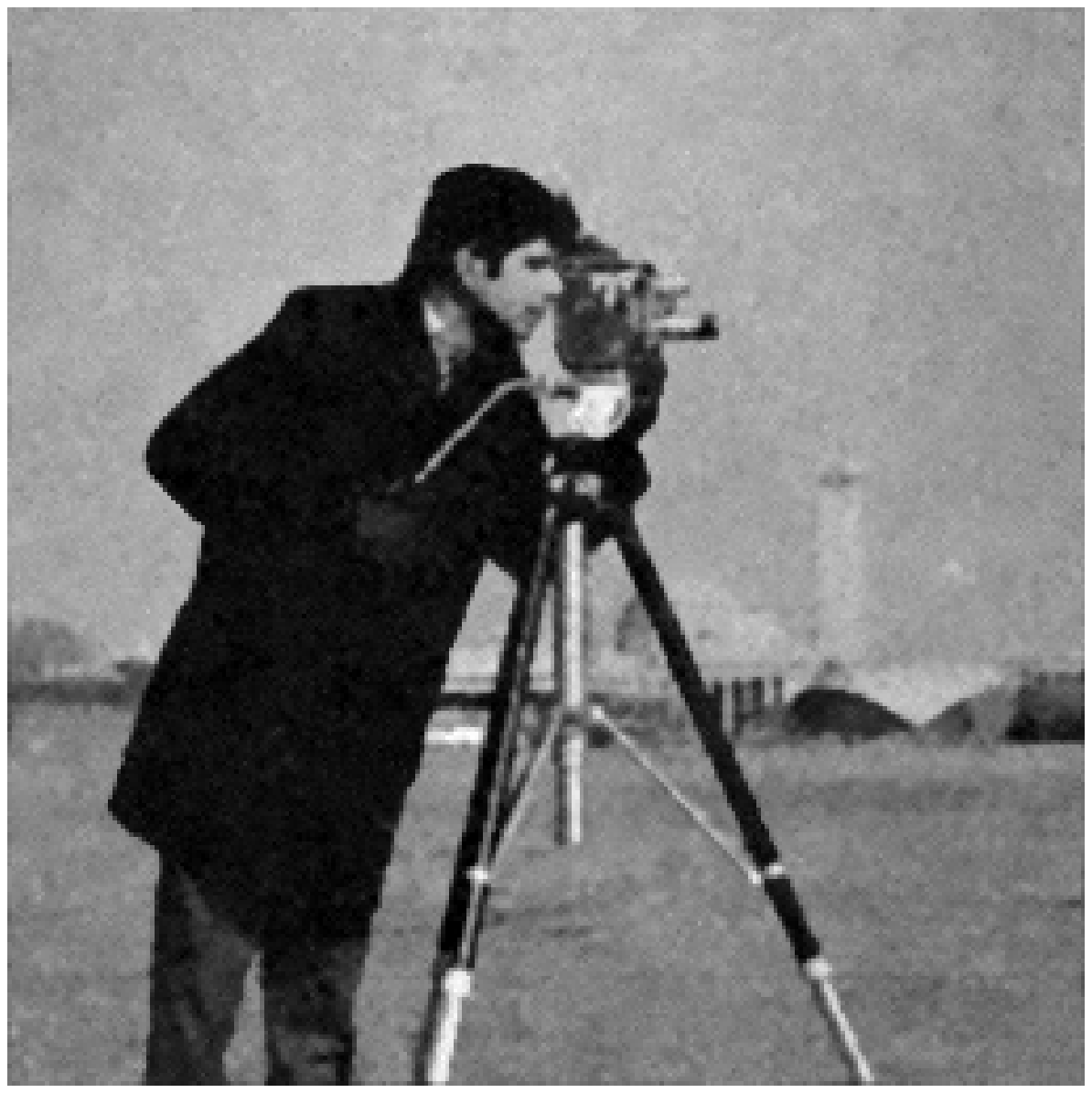}} 
\caption{Denoising results using standard bilateral filter (SBF) and the proposed weighted bilateral filter (WBF). For reproducibility, we used the builtin \textit{Camerman} image that comes with Matlab. 
We tuned the parameters of SBF and WBF to get the optimal PSNR in either case. The optimal ($\sigma_s,\sigma_r$) setting is indicated in the caption. 
} 
\label{Images}
\end{figure}

In Table \ref{table2}, the denoising performance of the proposed weighted filter at different noise level is compared with the standard and the robust bilateral filter, as well the patch-based non-local means filter \cite{Buades2005}. We have used $L=1$ in \eqref{box} for all the experiments.  
To have a fair comparison for a given image and noise level, we independently tuned $\sigma_s$ and $\sigma_r$ to optimize the PSNRs of the respective filters. We also tuned the parameters of NLM to get the optimal PSNR at each noise level. 
As remarked earlier,  the robust filter starts to perform better than the standard filter beyond a certain noise level ($\sigma \approx 20$). Notice that the PSNR obtained using the optimally-weighted filters is consistently higher than  that of the constituent filters at all noise levels; the improvement is often as large as $1$ dB. On the other hand, the improvement in PSNR over the standard bilateral filter is often as high as $10$ dB. This does not come as a surprise since it is well-known that the bilateral filter has a lot of scope of improvement at high-noise levels \cite{Book2009}.  
However, what does come as a surprise is that proposed filter is competitive with the non-local means filter \cite{Buades2005} that uses patches (groups of neighboring pixels) instead of single pixels for denoising.
For a visual comparison, the results of a particular denoising experiment are shown in Figure \ref{Images}. Notice that the denoised image obtained using the proposed filter looks much more sharp than that obtained using the standard filter, and has a significantly higher PSNR. Also, notice that the noise in the background is much less in (d) compared to (c).
In Table \ref{table3}, we compare the run times of the direct and the fast implementation of the proposed filter for different parameter settings. 
%We note that the main computation in either implementation is that of computing the standard and the robust bilateral filters, along with their SURE estimates (the cost of computing %the optimal weights by solving the $2\times 2$ linear system is negligible in comparison). 
Notice that the fast implementation is significantly faster than the direct implementation. 

\section{Conclusion}
\label{sec:conclusion}

We demonstrated how a simple pre-processing step can substantially improve the denoising performance of the bilateral filter.  
To consistently get the best of the standard and the pre-processed filter at all noise levels, we proposed to optimally weight them using Stein's unbiased estimate of 
the MSE. The optimal weights were found by solving a small linear system. 
A fast algorithm for implementing the optimally-weighted filters was also described. 
We reported visual and PSNR results on test images which confirmed the improvement over the original bilateral filter. 
An interesting finding was that the weighted bilateral filter is competitive with the non-local means filter. 

\vfill\pagebreak

\bibliographystyle{IEEEbib}

\end{document}